%% file: main.tex
\documentclass[twoside,11pt]{article}

\usepackage{automl2019}
\usepackage[utf8]{inputenc}
\usepackage{microtype}
\usepackage{graphicx}
\usepackage{amsmath}
\usepackage{subfigure}
\usepackage[bf]{caption} 
\usepackage{tabularx,booktabs} % for professional tables
\usepackage{todonotes}
\usepackage{hyperref}
\usepackage{multicol}
\usepackage{xpatch}

\usepackage{algorithm}
\usepackage{algorithmic}
\makeatletter
\g@addto@macro\normalsize{\setlength\abovedisplayskip{6pt}}
\makeatother

\makeatletter
\g@addto@macro\normalsize{\setlength\belowdisplayskip{6pt}}
\makeatother

 \jmlrheading{H.S. Behl, A.G. Baydin, and P.H.S. Torr}
\ShortHeadings{Alpha MAML: Adaptive Model-Agnostic Meta-Learning}{Behl, Baydin and Torr}
\firstpageno{1}

\begin{document}

\title{Alpha MAML: Adaptive Model-Agnostic Meta-Learning}

 \author{\name Harkirat Singh Behl \email harkirat@robots.ox.ac.uk \\ \name Atılım Güneş Baydin \email gunes@robots.ox.ac.uk \\ \name Philip H.S. Torr \email phst@robots.ox.ac.uk \\ \addr University of Oxford}

\maketitle
% \vspace{-1.2cm}
\begin{abstract}%   <- trailing '%' for backward compatibility of .sty file
Model-agnostic meta-learning (MAML) is a meta-learning technique to train a model on a multitude of learning tasks in a way that primes the model for few-shot learning of new tasks. The MAML algorithm performs well on few-shot learning problems in classification, regression, and fine-tuning of policy gradients in reinforcement learning, but comes with the need for costly hyperparameter tuning for training stability. We address this shortcoming by introducing an extension to MAML, called Alpha MAML, to incorporate an online hyperparameter adaptation scheme that eliminates the need to tune meta-learning and learning rates. Our results with the Omniglot database demonstrate a substantial reduction in the need to tune MAML training hyperparameters and improvement to training stability with less sensitivity to hyperparameter choice.
\end{abstract}

\input{sections/introduction.tex}
\input{sections/related_work.tex}
\input{sections/maml.tex}
\input{sections/alpha_maml.tex}
\input{sections/experiments.tex}
\input{sections/conclusion.tex}

\vskip 0.2in
\bibliography{main}
\bibliographystyle{automl2019}

\newpage
\appendix
\input{sections/appendix.tex}

\end{document}

%% file: sections/introduction.tex
%!TEX root = ../main.tex
\section{Introduction}
% \vspace{-0.1in}
Meta-learning---or ``learning to learn''---concerns machine learning models that can improve their learning quality by altering aspects of the learning process such as the model architecture, optimization rules, initialization, or learning hyperparameters \citep{thrun2012learning,schmidhuber1987evolutionary,hochreiter2001learning}. An important application of meta-learning is in few-shot learning problems \citep{NIPS2016_6385, DBLP:journals/corr/abs-1812-01397}, where one is concerned with developing methods able to learn new concepts from one or only a few instances \citep{lake2015human}. % mimicking the human ability of generalizing successfully from limited examples.
In this paper we focus on the state-of-the-art \emph{model-agnostic meta-learning} (MAML) \citep{finn2017model} method, which is a conceptually simple and general algorithm that has been shown to outperform existing approaches in tasks including few-shot image classification and few-shot adaptation in reinforcement learning \citep{antoniou2018how}. %Learning a new concept with stochastic gradient descent involves three important components, namely, initialization, which direction to move in (the negative gradient), and how much to move. In meta-learning, the direction can come from the gradients available from the data for the new concept.
MAML aims to solve the few-shot learning problem by being just few gradient descent steps away from any new concepts, doing so by making the assumption that learning a new concept will just involve few parameter updates (Algorithm~\ref{algo:maml}). In other words, MAML is based on learning an initial representation that can be efficiently fine-tuned for new tasks in a few steps.

The generality of MAML comes with the difficulty of choosing hyperparameters to achieve stable training in practice \citep{antoniou2018how}. MAML has two important hyper-parameters, namely the learning rate $\alpha$ and the meta-learning rate $\beta$, thus increasing any hyperparameter grid search computation by an order, and making it significantly more time and resource consuming than comparable methods. Another complication to this problem is the fact that it is currently not established whether the technique can benefit from a conventional decaying schedule for the inner learning rate $\alpha$. Furthermore, a good value of $\alpha$ in MAML is even more important than for any conventional stochastic gradient descent (SGD) optimization, because only a handful of samples are available in the few-shot learning case. This has significant consequences, making it difficult to scale this algorithm to problems bigger than toy scales, due to the difficulty in assessing whether MAML is not suitable for a complex task or whether the hyperparameters are not sufficiently tuned.

In this paper, we provide a conceptually simple solution to this problem, by introducing an extension of the MAML algorithm to incorporate adaptive tuning of both the learning rate $\alpha$ and the meta-learning rate $\beta$. Our aim is to make it possible to use MAML without or with significantly less parameter tuning, and thus to reduce the need for grid search. We also aim to make the algorithm converge in fewer iterations. The solution we propose is based on the hypergradient descent (HD) algorithm \citep{baydin2018hypergradient}, which automatically updates a learning rate by performing gradient descent on the learning rate alongside original optimization steps. The proposed algorithm does not need any extra gradient computations, and just involves storing the gradients from the previous optimization step.

% In normal SGD, you have one parameter alpha $\alpha$, and people have come up with a lot of ways to tune this parameter. Naive solution is to do a grid search over the hyper-parameters.
% Another approach which is most common is using adaptive algorithms. Because you are not sure about the hyper-parameter value that you chose, you would like to adapt the value of alpha as the training progresses. Also because a decaying schedule for alpha is preferred. Some of the common techniques involve ??

% Bengio calls it the single most important hyperparameter.
% All variants of SGD have been proposed around this.

%In Section~\ref{sec:related_work} we cover related work. In Sections \ref{sec:maml} and \ref{sec:alpha_maml}, we introduce the Alpha MAML algorithm in reference to the original MAML algorithm. In Section~\ref{sec:experiments} we present adaptive learning rate tuning results with the Omniglot dataset, followed by conclusions in Section~\ref{sec:conclusion}.

%% file: sections/related_work.tex
%!TEX root = ../main.tex
% \vspace{-0.1in}
\section{Related Work}
\label{sec:related_work}
% \vspace{-0.1in}
Our work is primarily related with the subfields of hyperparameter optimization and meta-learning. In hyperparameter optimization one typically uses parallel runs to populate a selected grid of hyperparameter values (e.g., a range of learning rates), or use more advanced techniques such as Bayesian optimization \citep{snoek2012practical} and model-based approaches \citep{bergstra2013making,hutter2013evaluation}. An interesting line of research, which also inspired our approach in this paper, is to use gradient-based optimization for the tuning of hyperparameters \citep{bengio2000gradient}. Recent work in this area include reversible learning \citep{maclaurin2015gradient}, which allows gradient-based optimization of hyperparameters through a training run consisting of multiple iterations, and hypergradient descent \citep{baydin2018hypergradient}, which achieves a similar optimization in an online, per-gradient-update, fashion.
% \HB{Fix}

Meta-learning is often referred to as ``learning to learn'' \citep{thrun2012learning}, meaning a learning procedure (most of the time gradient-based) is able to improve aspects of the learning process itself, such as the optimizer, hyperparameters like the learning rate, and initializations. In this sense, the ``meta'' concept of meta-learning has aspects in common with hyperparameter optimization. %Meta-learning is usually expressed as consisting of task- and meta-levels of learning, where task-level refers to the fast acquisition of task-specific knowledge, and meta-level refers to the slow learning of across-task knowledge that allows generalization across tasks. 
The MAML model \citep{finn2017model} on which we base our method,  relies on meta-optimization through gradient descent in a model-agnostic way. Another recent method, Meta-SGD \citep{meta_sgd}, performs online optimization of a per-parameter learning rate vector $\alpha$, to which the authors refer as learning both the learning rate and update direction (because of the per-parameter nature being able to modify direction), and model parameters $\theta$, using a single hyper-learning rate $\beta$. Our work differs from Meta-SGD as we perform simultaneous online optimization of both MAML learning rates $\alpha$ and $\beta$, which are both scalars.

%% file: sections/maml.tex
\section{Model-Agnostic Meta-Learning (MAML)}
% \vspace{-0.1in}
\label{sec:maml}

The MAML algorithm, given model parameters $\theta$, aims to adapt to a new task $\mathcal{T}_t$ with SGD:
\begin{equation} \label{eq:inner-op-MAML}
    \theta_t^\prime = \theta - \alpha \nabla _{\theta} \mathcal{L}_{\mathcal{T}_{train(t)}}(f_{\theta})\;,
\end{equation}

where $t$ is the task number and $\alpha$ is the learning rate. ${T}_{train(t)}$ and ${T}_{test(t)}$ denote the training and test set within task $t$. The tasks are sampled from a defined $p(\mathcal{T})$. The meta-objective is:
\begin{equation} \label{eq:meta-op-MAML}
    \min_{\theta} \mathcal{L}_{\mathcal{T}_{test(t)}}(f_{\theta_t^\prime}) =   \mathcal{L}_{\mathcal{T}_{test(t)}}(f_{\theta - \alpha \nabla _{\theta} \mathcal{L}_{\mathcal{T}_{train(t)}}(f_{\theta})})
\end{equation}

The model aims to optimize the parameters $\theta$ such that with just one SGD step it can adapt to the new task. For the optimization in Eq. \ref{eq:meta-op-MAML}, this looks as follows:
\begin{equation} \label{eq:outer-op-MAML}
    \theta_t = \theta - \beta \nabla _{\theta} \mathcal{L}_{\mathcal{T}_{test(t)}}(f_{\theta_t^\prime})\;, 
\end{equation}
where $\beta$ is the meta step size. This gives an algorithm that learns an initialization of $\theta$ that is useful for being adapted to new tasks efficiently with a small number of iterations. An important advantage is that this is achieved without making any assumptions on the form of the model. Another advantage is that the meta-learner, introduced in this way based on conventional gradient descent, does not introduce extra model parameters to learn as in model-based approaches in meta-learning. The full algorithm is shown in Algorithm~\ref{algo:maml}. It can be seen that unlike usual SGD which has only one learning rate, MAML has two learning rates $\alpha$ and $\beta$, which require time-consuming hyperparameter tuning. %In the next section we describe our method, Alpha-MAML, which addresses this issue.

%% file: sections/alpha_maml.tex
%!TEX root = ../main.tex
% \vspace{-0.1in}
\section{Alpha MAML}
% \vspace{-0.in}
\label{sec:alpha_maml}
\begin{table}[]
	% \vspace{-0.4in}
% \begin{multicols*}{2}
\vspace{-5mm}
	\begin{algorithm}[H]
		\fontsize{8pt}{8pt}\selectfont
		\caption{{\small MAML}}
		\label{algo:maml}
		\begin{algorithmic}
		\STATE {\bfseries Input:} $p(\mathcal{T}):$ distribution over tasks.
		\STATE {\bfseries Input:} $\alpha, \beta:$ learning rates
		\STATE randomly initialize $\theta$.
		\WHILE {not done}
		\STATE Sample batch of tasks $\mathcal{T}_t \sim p(\mathcal{T})$
		\FOR {all $\mathcal{T}_t$ }
		\STATE Evaluate $\nabla _{\theta}\mathcal{L}_{\mathcal{T}_{train(t)}}(f_{\theta})$ with respect to $K$ examples.
		\STATE Compute adapted parameters with gradient descent: $\theta_t^\prime = \theta - \alpha \nabla _{\theta} \mathcal{L}_{\mathcal{T}_{train(t)}}(f_{\theta})$
		\ENDFOR
		\STATE Update $\theta = \theta - \beta \nabla _{\theta} \sum_{\mathcal{T}_t \sim p(\mathcal{T})} \mathcal{L}_{\mathcal{T}_{test(t)}}(f_{\theta_t^\prime})$
		\ENDWHILE
		\end{algorithmic}
	\end{algorithm}
% \columnbreak
\vspace{-2mm}
	\begin{algorithm}[H]
		\fontsize{8pt}{8pt}\selectfont
		\caption{{\small Alpha MAML}}
		\label{algo:alpha_maml}
		\begin{algorithmic}
			\STATE {\bfseries Input:} $p(\mathcal{T}):$ distribution over tasks.
			\STATE {\bfseries Input:} $\alpha_0, \beta_0:$ initial learning rates
			\STATE {\bfseries Input:} $\alpha_\text{hyperlr}, \beta_\text{hyperlr}:$ hypergradient learning rates
			\STATE randomly initialize $\theta$.
			\WHILE {not done}
			\STATE Sample batch of tasks $\mathcal{T}_t \sim p(\mathcal{T})$
			\FOR {all $\mathcal{T}_t$ }
			\STATE Evaluate $\nabla _{\theta}\mathcal{L}_{\mathcal{T}_{train(t)}}(f_{\theta})$ with respect to $K$ examples.
			\STATE Compute adapted parameters with gradient descent: $\theta_t^\prime = \theta - \alpha_i \nabla _{\theta} \mathcal{L}_{\mathcal{T}_{train(t)}}(f_{\theta})$
			\ENDFOR
			\STATE $\alpha_{i+1}=\alpha_{i} + \alpha_\text{hyperlr} \sum_{\mathcal{T}_t \sim p(\mathcal{T})} \nabla _{\theta_{t}^\prime}\mathcal{L}_{\mathcal{T}_{test(t)}}(f_{\theta_{t}^\prime}).\nabla _{\theta_{i-1}} \mathcal{L}_{\mathcal{T}_{train(t)}}(f_{\theta_{i-1}})$
			\STATE $\beta_{i} = \beta_{i-1} + \beta_\text{hyperlr}\sum_{\mathcal{T}_t \sim p(\mathcal{T})} \nabla _{\theta_{i-1}} \mathcal{L}_{\mathcal{T}_{test(t)}}(f_{\theta_t^\prime}).\nabla _{\theta_{i-2}} \mathcal{L}_{\mathcal{T}_{test(i-1)}}(f_{\theta_{t}^\prime})$
			\STATE $\theta_i = \theta_{i-1} - \beta_i  \sum_{\mathcal{T}_t \sim p(\mathcal{T})} \nabla _{\theta_{i-1}}\mathcal{L}_{\mathcal{T}_{test(t)}}(f_{\theta_t^\prime})$
			\ENDWHILE
		\end{algorithmic}
	\end{algorithm}
% \end{multicols*}
\vspace{-6mm}
\end{table}

Eq.~\ref{eq:inner-op-MAML} is regular gradient descent for adapting to the task $t$, with $\alpha$ being the task-level learning rate. Similarly, Eq.~\ref{eq:outer-op-MAML} is regular gradient descent for the meta update, with $\beta$ being the meta-learning rate.
In addition to the update rules in Equations~\ref{eq:inner-op-MAML} and \ref{eq:outer-op-MAML}, we would like to derive update rules for the learning rates $\alpha$ and $\beta$ as well. Our algorithm, Alpha MAML, can simply be written in four update equations as shown in the following derivation. Here $i$ is the iteration number. We first derive the algorithm for the simpler case when each batch has only one task, thus the iteration number is same as the task number $t=i$.

Firstly, our goal is to update the value of $\alpha$ towards the optimum value $\alpha_i^*$ that minimizes the value of the meta objective Eq.~\ref{eq:meta-op-MAML} in the next iteration. However, we have not computed $\theta_i^\prime$ yet. If we assume that the optimal value of $\alpha$ does not change much across iterations, we can estimate it by $\alpha_{i-1}^*$. Therefore we perform one gradient descent step over the previous value of learning rate $\alpha_{i-1}$, with gradient:
\begin{equation}\label{eq:der-alpha}
\begin{aligned}
    \frac{\partial \mathcal{L}_{\mathcal{T}_{test(i-1)}}(f_{\theta_{i-1}^\prime})}{\partial \alpha} &= \frac{\partial \mathcal{L}_{\mathcal{T}_{test(i-1)}}(f_{\theta_{i-1}^\prime})}{\partial \theta_{i-1}^\prime}^T \frac{\partial \theta_{i-1}^\prime}{\partial \alpha} \\
    &= {\nabla _{\theta_{i-1}^\prime} \mathcal{L}_{\mathcal{T}_{test(i-1)}}(f_{\theta_{i-1}^\prime})}^T\\
    &\;\;\;\;\;\;\;\;\;\frac{\partial ( \theta_{i-2} - \alpha_{i-1} \nabla _{\theta_{i-2}} \mathcal{L}_{\mathcal{T}_{train(i-1)}}(f_{\theta_{i-2}}))} {\partial \alpha} \\
    &= \nabla _{\theta_{i-1}^\prime} \mathcal{L}_{\mathcal{T}_{test(i-1)}}(f_{\theta_{i-1}^\prime}) . ( - \nabla _{\theta_{i-2}} \mathcal{L}_{\mathcal{T}_{train(i-1)}}(f_{\theta_{i-2}}))
\end{aligned}
\end{equation}

We can estimate $\alpha_{i-1}^*$ as follows:
\begin{equation}
\begin{aligned}
    &\alpha_{i} = \alpha_{i-1} + \alpha_\text{hyperlr} \nabla _{\theta_{i-1}^\prime} \mathcal{L}_{\mathcal{T}_{test(i-1)}}(f_{\theta_{i-1}^\prime})\,.
    \nabla _{\theta_{i-2}} \mathcal{L}_{\mathcal{T}_{train(i-1)}}(f_{\theta_{i-2}})\;,
\end{aligned}
\end{equation} %\label{eq:inner-op-alpha-MAML}
where $\alpha_\text{hyperlr}$ is the hyper learning rate for $\alpha$.

Secondly, we also want to derive an update rule for the meta learning rate $\beta$. Similar to $\alpha$, we would like to update $\beta$ towards its optimal value $\beta_i^*$ that minimizes the value of the objective $\mathcal{L}_{\mathcal{T}_{test(i)}}(f_{\theta_i^\prime})$ in the next iteration, making an assumption that the optimal value of $\beta$ does not change much between two consecutive iterations. For this, we compute:
\begin{equation}\label{eq:der-beta}
\begin{aligned}
    \frac{\partial \mathcal{L}_{\mathcal{T}_{test(i)}}(f_{\theta_{i}^\prime})}{\partial \beta} 
    &= \frac{\partial \mathcal{L}_{\mathcal{T}_{test(i)}}(f_{\theta_{i}^\prime})}{\partial \theta_{i-1}}. \frac{\partial \theta_{i-1}}{\partial \beta} \\
    &= \nabla _{\theta_{i-1}} \mathcal{L}_{\mathcal{T}_{test(i)}}(f_{\theta_i^\prime}) . (-\nabla _{\theta_{i-2}} \mathcal{L}_{\mathcal{T}_{test(i-1)}}(f_{\theta_{i-1}^\prime}))\\
\end{aligned}
\end{equation}

We can estimate $\beta_{i-1}^*$ as follows:
\begin{equation}%\label{eq:inner-op-alpha-MAML}
    \beta_{i} = \beta_{i-1} + \beta_\text{hyperlr} \nabla _{\theta_{i-1}} \mathcal{L}_{\mathcal{T}_{test(i)}}(f_{\theta_i^\prime}). \nabla _{\theta_{i-2}} \mathcal{L}_{\mathcal{T}_{test(i-1)}}(f_{\theta_{i-1}^\prime})\;,
\end{equation}

where $\beta_\text{hyperlr}$ is the hyper learning rate for $\beta$. It is important to note that in Eq.\ref{eq:der-alpha} and Eq.\ref{eq:der-beta}, the gradients are computed with respect to the loss on the test set of a batch. This is done in accordance with the meta-objective.
%%%%%%%%%%%%%%%%%%%%%%%%%%%%%%%%%%%%%%%%%%%%%%%%%%%%%%%%%%%%%%%%%%%%%
%%%%%%%%%%%%%%%%%%%%%%%%%%%%%%%%%%%%%%%%%%%%%%%%%%%%%%%%%%%%%%%%%%%%%
%%%%%%%%%% Alpha MAML %%%%%%%%%%%%%%%%%%%%%
%%%%%%%%%%%%%%%%%%%%%%%%%%%%%%%%%%%%%%%%%%%%%%%%%%%%%%%%%%%%%%%%%%%%%
The final Alpha MAML algorithm can thus be written down into just 4 update equations:

\begin{equation} %\label{eq:Alpha-MAML}
\begin{alignedat}{2}
& \theta_i^\prime = \theta_{i-1} - \alpha_i \nabla _{\theta_{i-1}} \mathcal{L}_{\mathcal{T}_{train(i)}}(f_{\theta_{i-1}}) \\
& \alpha_{i+1} = \alpha_{i} + \alpha_\text{hyperlr} \nabla _{\theta_{i}^\prime} \mathcal{L}_{\mathcal{T}_{test(i)}}(f_{\theta_{i}^\prime}) . \nabla _{\theta_{i-1}} \mathcal{L}_{\mathcal{T}_{train(i)}}(f_{\theta_{i-1}}) \\
& \beta_{i} = \beta_{i-1} + \beta_\text{hyperlr} \nabla _{\theta_{i-1}} \mathcal{L}_{\mathcal{T}_{test(i)}}(f_{\theta_i^\prime}) .\nabla _{\theta_{i-2}} \mathcal{L}_{\mathcal{T}_{test(i-1)}}(f_{\theta_{i-1}^\prime}) \\
& \theta_i = \theta_{i-1} - \beta_i \nabla _{\theta_{i-1}} \mathcal{L}_{\mathcal{T}_{test(i)}}(f_{\theta_i^\prime}) 
\end{alignedat}
\end{equation}

It can be seen that no extra gradient needs to be computed, as the gradients from the last iteration can be used, requiring only the extra memory storage of the gradient from the previous iteration.
The full algorithm with the more general case of multiple tasks in one batch is shown in Algorithm \ref{algo:alpha_maml}. The derivation for this case is shown in the appendix \ref{appendix_derivation}.
% \item multiple steps within each task.
% \item and multiple examples within each task (k-shot).
% \item multiplicative rule for hypergradient learning rate.
% \item works with Adam etc.
% \item works with How to Train your MAML \cite{antoniou2018how}
% \end{enumerate}

%% file: sections/experiments.tex
%!TEX root = ../main.tex
% \vspace{-0.1in}
\section{Experiments}
% \vspace{-0.1in}
\label{sec:experiments}
To evaluate the performance of Alpha MAML in comparison to MAML, we perform experiments on the few-shot image recognition task on the Omniglot dataset \citep{lake2011one}, which is commonly used in related work \citep{finn2017model, ravi2016optimization, NIPS2017_6996, NIPS2016_6385}. We keep the experimentation configuration the same as the original MAML work, for a fair comparison.
Omniglot comprises 20 instances (drawn by 20 different people) of 1,623 characters from 50 alphabets. We follow the N-way Omniglot task setup, which was introduced by \citet{NIPS2016_6385}, and also used in MAML \citep{finn2017model}. The N-way classification task is set up as follows: the model is shown K instances from N unseen classes, and then evaluated on some other instances from these N classes. To keep the procedure the same as MAML, we also randomly select 1,200 characters for training, and use the rest for testing.
The network architecture follows the architecture used by \cite{finn2017model}, and more implementation details are in the appendix \ref{appendix_implement}.

%%%%%%%%%%%%%%%%%%%%%%%%%%%%%%%%%%%%%%
\begin{figure}[]
	\vspace{-0.1in}
	\centering
	\captionsetup{format=plain}
	\includegraphics[width=0.32\textwidth]{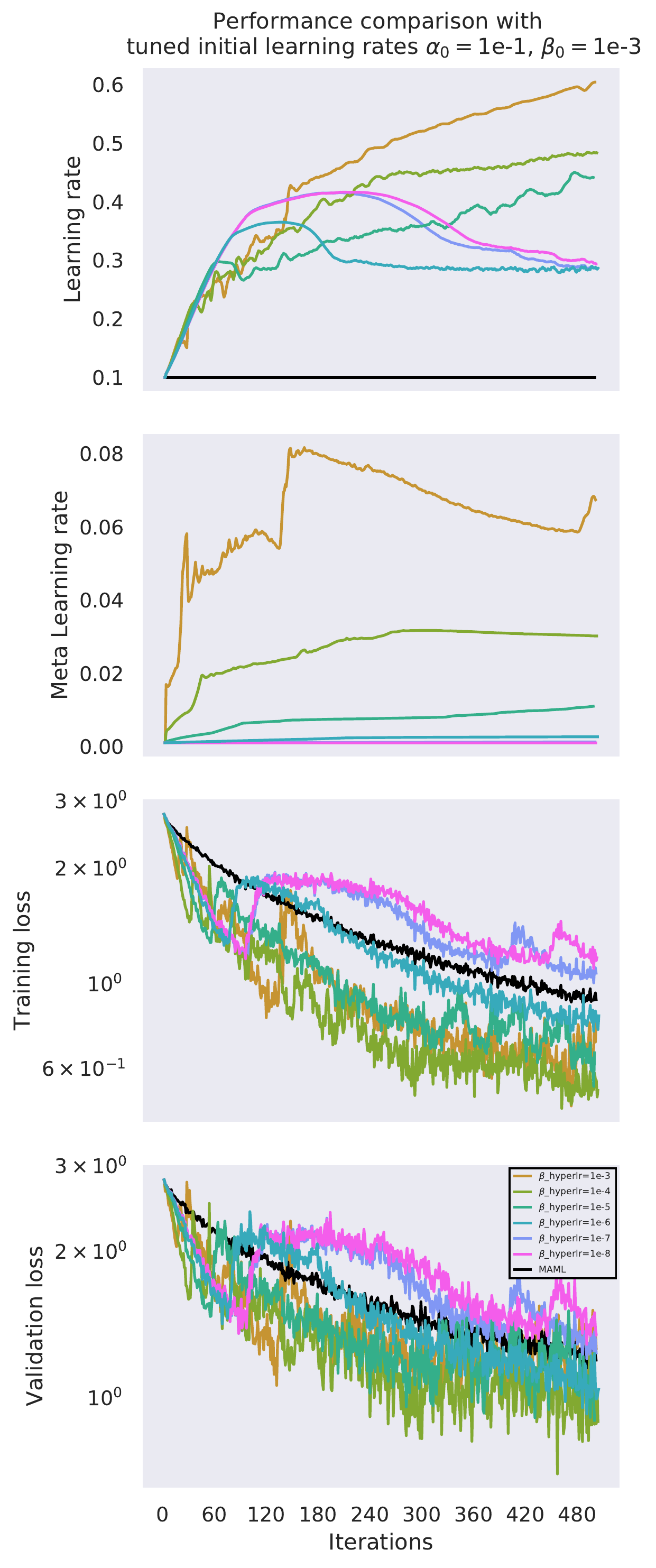}
	%\hspace*{\fill} % separation between the subfigures
	\includegraphics[width=0.32\textwidth]{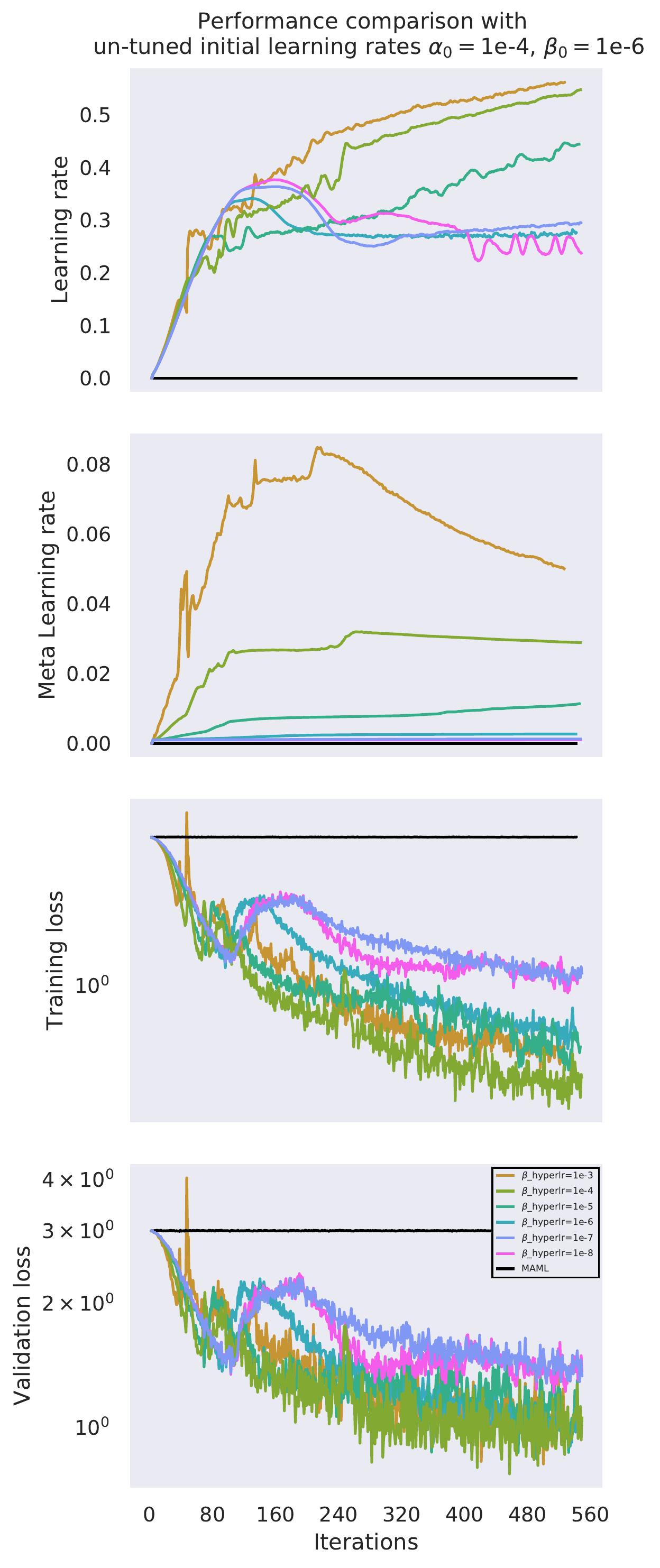}
	%\hspace*{\fill} % separation between the subfigures
	\includegraphics[width=0.32\textwidth]{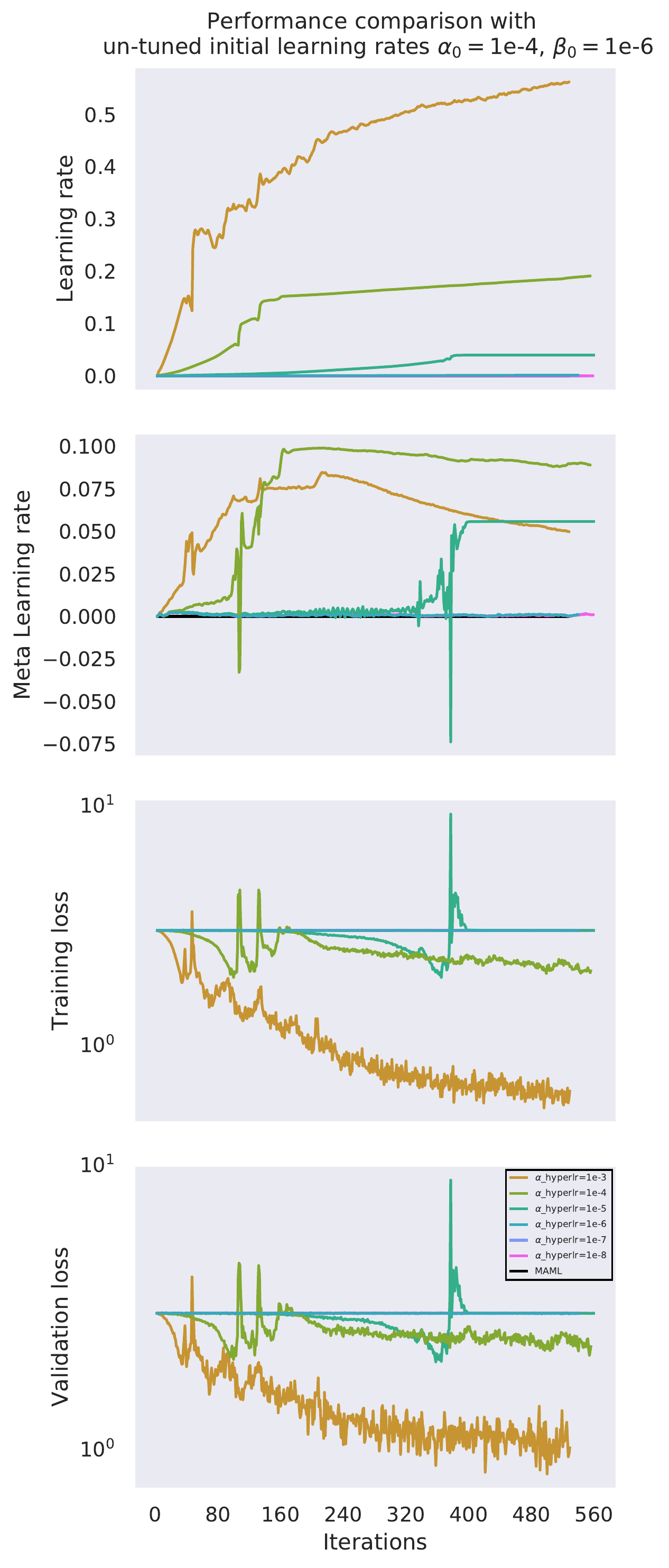}
	%\hspace*{\fill} % separation between the subfigures
	\vspace{-0.1in}
	\caption{\footnotesize \textbf{Convergence comparison of Alpha-MAML and MAML, with and without tuned initial learning rate hyper-parameters $\alpha_{0}$ and $\beta_{0}$}. Columns: \textit{left:} tuned initial learning rates; \textit{middle:} untuned initial learning rates, with varying $\beta_\text{hyperlr}$ for Alpha-MAML; \textit{right:} untuned initial learning rates, with varying $\alpha_\text{hyperlr}$ for Alpha-MAML. Rows: \textit{first:} evolution of the learning rate $\alpha$; \textit{second:} evolution of the meta learning rate $\beta$; \textit{third:} training loss; \textit{fourth:} validation loss.} 
	\label{fig:training}
	\vspace{-4mm}
\end{figure}
%%%%%%%%%%%%%%%%%%%%%%%%%%%%%%%%%%%%%%
\paragraph{Behaviour of Alpha MAML vs MAML:}
\label{sec:training}

Here we choose a good case (where the MAML user picked a good pair of $\alpha_{0}$ and $\beta_{0}$, i.e., the tuned case) and a bad case (where the user picked a bad pair of $\alpha_{0}$ and $\beta_{0}$, i.e., an untuned case), and plot the evolution of $\alpha_i$ and $\beta_i$ as a function of iterations, also showing the training and validation losses during training.
%Note that for regular MAML the learning and meta-learning rates are denoted $\alpha$ and $\beta$, whereas in Alpha MAML, we supply the initial values of these denoted as $\alpha_0$ and $\beta_0$ and let them evolve during the training under the hypergradient descent rules derived in Section~\ref{sec:alpha_maml}.
Figure~\ref{fig:training} shows the evolution of learning rate $\alpha_i$, meta-learning rate $\beta_i$, and the training and validation losses. 
%We also show the case for regular MAML with these good and badly picked $\alpha_{0}$ and $\beta_{0}$ values.
Results show that even the badly picked $\alpha_{0}$ and $\beta_{0}$ values can be automatically tuned by the online learning rate adaptation scheme in Alpha MAML, tuned in the sense that the algorithm does the necessary adjustments to $\alpha_i$ and $\beta_i$ in each iteration to achieve a loss similar to the good case. This can be seen in Figure~\ref{fig:training} \textit{middle} and \textit{right} columns. %For regular MAML, they are just $\alpha$ and $\beta$, because they are constant during the whole run.

\paragraph{Insensitivity with respect to hyperparameter choices:}
\setlength{\tabcolsep}{1pt}
\begin{figure}[]
	\vspace{-0.2in}
	\vspace{-0.5\baselineskip}
	\begin{tabular}{m{3cm}cccc}
		\includegraphics[width=0.24\textwidth]{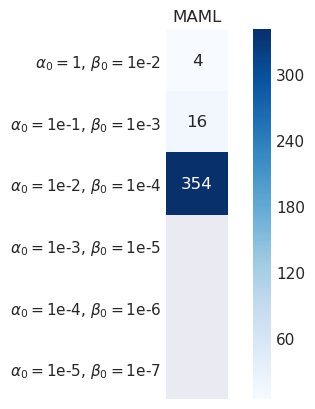}
		& 
		{
			\vspace{-\baselineskip} 
			\begin{tabular}{c} 
				\includegraphics[width=0.24\textwidth]{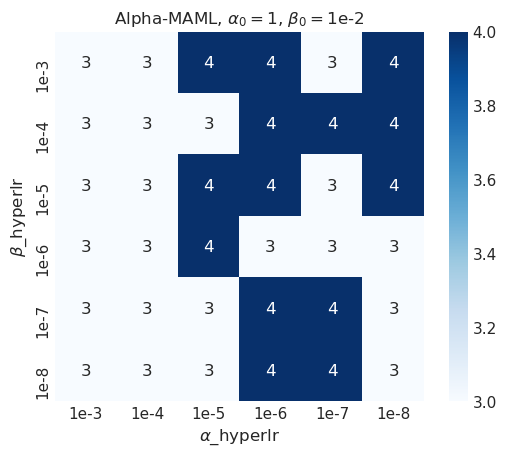} \\
				\includegraphics[width=0.24\textwidth]{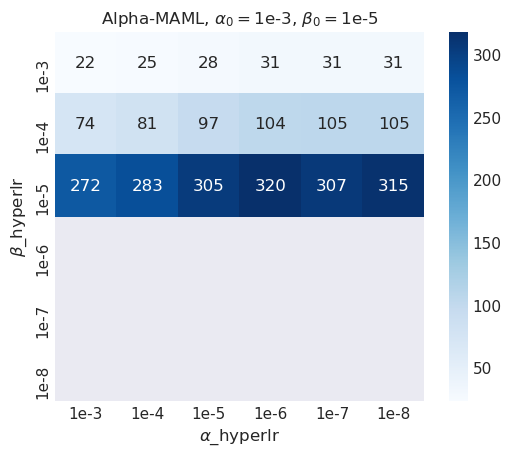}
			\end{tabular}
		} &
		\begin{tabular}{c}
			% \hline
			\includegraphics[width=0.24\textwidth]{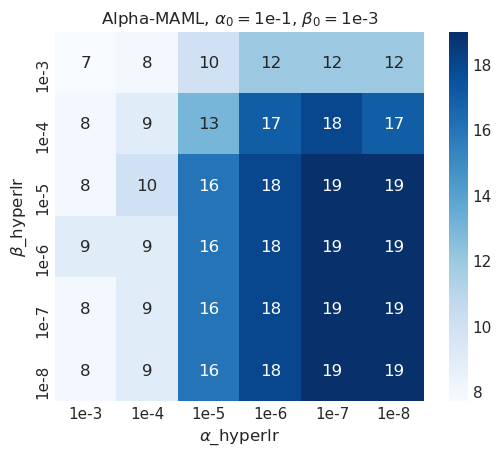} \\
			% \hline
			\includegraphics[width=0.24\textwidth]{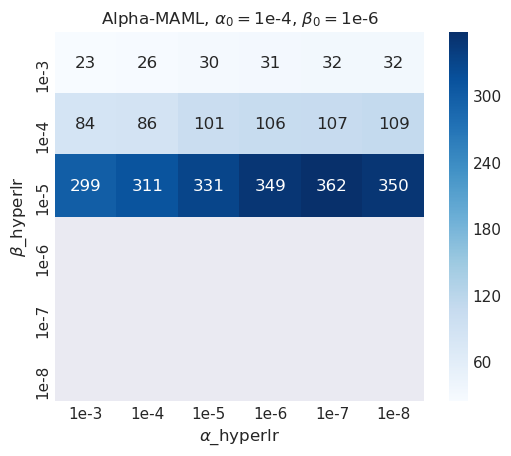}
		\end{tabular}
		&
		\begin{tabular}{c}
			% \hline
			\includegraphics[width=0.24\textwidth]{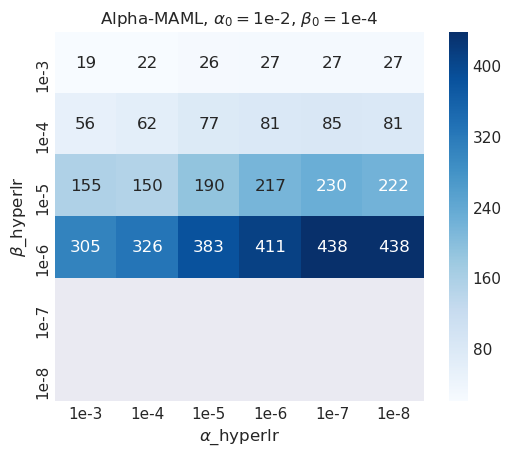} \\
			% \hline
			\includegraphics[width=0.24\textwidth]{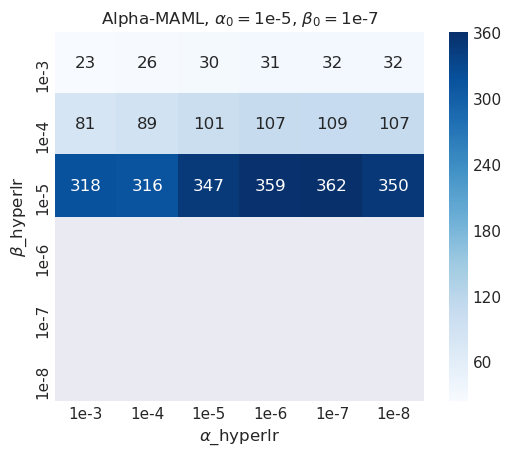}
		\end{tabular}
	\end{tabular}
\vspace{-0.1in}
	\caption{\footnotesize \textbf{Grid Search for selecting the hyper-parameters in MAML and Alpha-MAML: shows the number of iterations to converge to a training loss of 2.5 for Omniglot dataset.} Here the blank cells denote the cases where algorithm did not converge till 500 iterations. It can be seen that as we go far from the tuned hyper-parameter range, converge of Alpha-MAML is better than MAML. Specially for the case of $\alpha_{0}=1e-3,  \beta_{0}=1e-5$ Alpha-MAML converged for some cases of $\alpha_\text{hyperlr}$ and $\beta_\text{hyperlr}$, whereas MAML does not converge till 500 iterations. Same is the case with $\alpha_{0}=1e-4,  \beta_{0}=1e-6$, for half of the choices of $\alpha_\text{hyperlr}$ and $\beta_\text{hyperlr}$, Alpha-MAML converged but MAML does not converge in less than 500 iterations.}
	\label{fig:grid_search}
	\vspace{-4mm}
\end{figure}

Here we run a series of training experiments to study the effect of initial learning rate values on the number of iterations needed for the algorithms to reach a particular chosen loss threshold. Figures \ref{fig:grid_search} and \ref{fig:grid_search2} show this grid search for tuning the learning rate hyper-parameters. It can be seen that MAML shows very slow convergence for a range of initial learning rates, more specifically for $\alpha_{0}=1e-4,  \beta_{0}=1e-6$ and $\alpha_{0}=1e-5,  \beta_{0}=1e-7$. In comparison, Alpha MAML shows comparatively faster convergence for these initial learning rates also, for a wide range of $\alpha_\text{hyperlr}$ and $\beta_\text{hyperlr}$.
For the cases where MAML shows fast convergence, Alpha MAML also shows fast convergence for all values of $\alpha_\text{hyperlr}$ and $\beta_\text{hyperlr}$. This indicates that in practice no matter which values one chooses for the inital learning and meta-learning rates $\alpha_0$ and $\beta_0$, Alpha MAML always shows faster, or in the worst case the same, convergence as MAML. In other words, Alpha MAML is less sensitive to the hyperparameter choice as compared to MAML, and hence needs significantly less tuning.

% \newpage
% \subsection{Regression}
% \subsection{Reinforcement Learning}

%% file: sections/conclusion.tex
%!TEX root = ../main.tex
% \vspace{-0.1in}
\section{Conclusion}
\label{sec:conclusion}
% \vspace{-0.1in}
Meta-learning is an important and currently relevant approach with potential impact in solving hard problems in computer vision and reinforcement learning. The training instability of meta-learning algorithms as a function of hyperparameter choices is a known shortcoming and currently an active area of research. 
We have proposed an extension of the state-of-the-art MAML algorithm \citep{finn2017model} based on the application of the hypergradient descent technique \citep{baydin2018hypergradient} to make MAML training more robust to hyperparameter choices.
%Our technique, Alpha MAML, achieves the online learning rate adaptation of both regular- and meta-learning rates, using only the existing gradient information already available in the regular gradient-descent procedure. Our experiments with the Omniglot dataset demonstrate that the automated tuning of these hyperparameters constitutes a promising extension and can save valuable time and compute resources in practice.

%% file: sections/appendix.tex
%!TEX root = ../main.tex

%%%%%%%%%%%%%%%%%%%%%%%%%%%%%%%%%%%%%

\section{Appendix}
\label{sec:appendix}

We also derive the Alpha MAML update equations for the bigger case of multiple tasks in one batch as follows, where $i$ denotes the iteration number and $t$ is used to denote the task index:

% \begin{subequations} %\label{eq:Alpha-MAML}
\begin{equation}
\begin{alignedat}{2}
& \theta_t^\prime = \theta_{i-1} - \alpha_i \nabla _{\theta_{i-1}} \mathcal{L}_{\mathcal{T}_{train(t)}}(f_{\theta_{i-1}}) \\
& \alpha_{i+1} = \alpha_{i} + \alpha_\text{hyperlr} \sum_{\mathcal{T}_t \sim p(\mathcal{T})} \nabla _{\theta_{t}^\prime} \mathcal{L}_{\mathcal{T}_{test(t)}}(f_{\theta_{t}^\prime})\,.\nabla _{\theta_{i-1}} \mathcal{L}_{\mathcal{T}_{train(t)}}(f_{\theta_{i-1}}) \\
& \beta_{i} = \beta_{i-1} + \beta_\text{hyperlr} \sum_{\mathcal{T}_t \sim p(\mathcal{T})} \nabla _{\theta_{i-1}} \mathcal{L}_{\mathcal{T}_{test(t)}}(f_{\theta_t^\prime})\,.\nabla _{\theta_{i-2}} \mathcal{L}_{\mathcal{T}_{test(i-1)}}(f_{\theta_{t}^\prime}) \\
& \theta_i = \theta_{i-1} - \beta_i  \sum_{\mathcal{T}_t \sim p(\mathcal{T})} \nabla _{\theta_{i-1}}\mathcal{L}_{\mathcal{T}_{test(t)}}(f_{\theta_t^\prime}) \\
\end{alignedat}
\end{equation}
% \end{subequations}
\label{appendix_derivation}
In the update for $\beta$, the second gradient is the previous step's gradient. %%%%%%%%%%%%%%%%%%%%%%%%%%%%%%%%%%%%%

%%%%%%%%%%%%%%%%%%%%%%%%%%%%%%%%%%%%%%
\begin{figure*}
	\centering
	\includegraphics[width=0.32\textwidth]{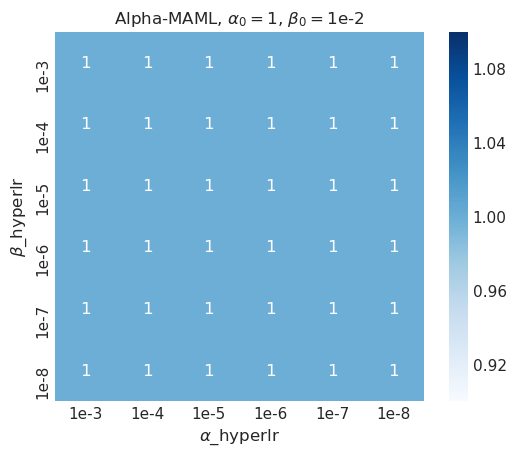}
	%\hspace*{\fill} % separation between the subfigures
	\includegraphics[width=0.32\textwidth]{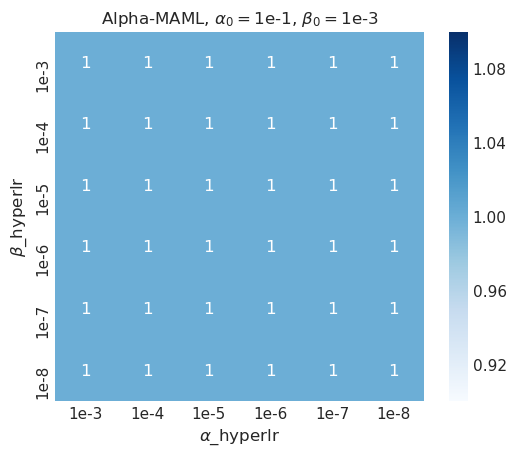}
	%\hspace*{\fill} % separation between the subfigures
	\includegraphics[width=0.32\textwidth]{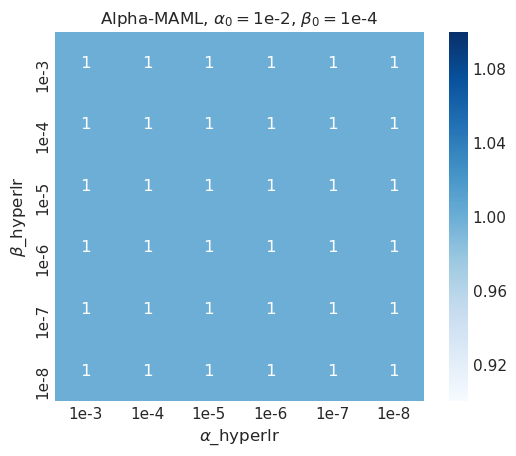}
	%\hspace*{\fill} % separation between the subfigures
	
	\includegraphics[width=0.32\textwidth]{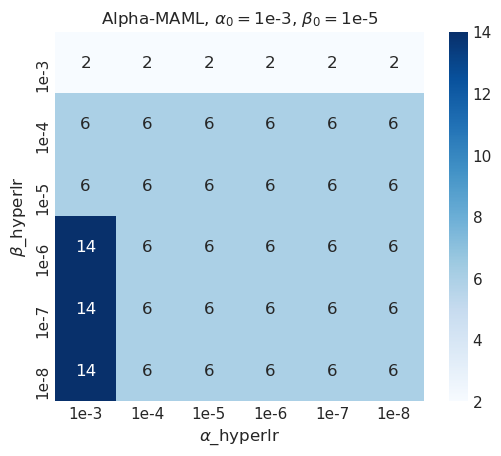}
	%\hspace*{\fill} % separation between the subfigures
	\includegraphics[width=0.32\textwidth]{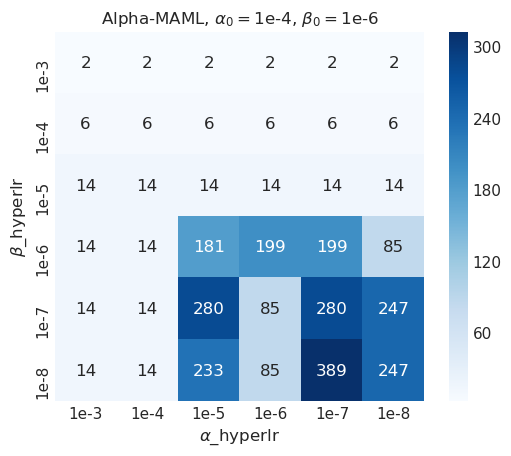}
	%\hspace*{\fill} % separation between the subfigures
	\includegraphics[width=0.32\textwidth]{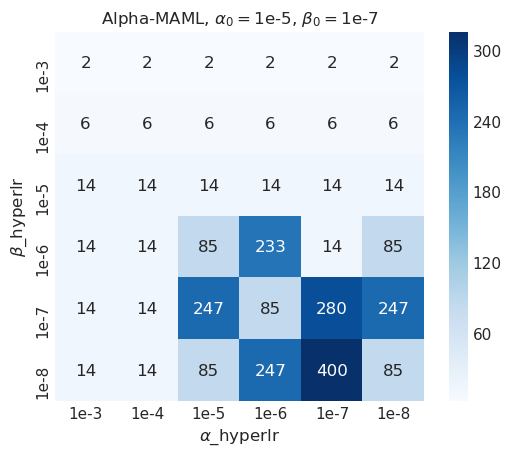}
	%\hspace*{\fill} % separation between the subfigures
	
	\includegraphics[width=0.32\textwidth]{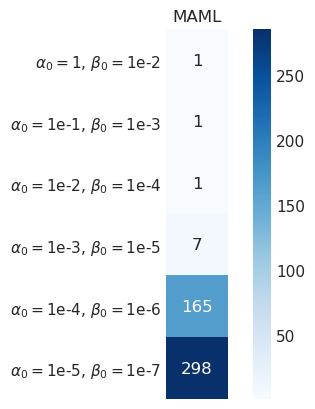}
	
	\caption{\textbf{Grid Search for selecting the hyper-parameters in MAML and Alpha-MAML: shows the number of iterations to converge to a training loss of 2.99 for Omniglot dataset.}  It can be seen that as we go far from the tuned hyper-parameter range, converge of Alpha-MAML is better than MAML. Specially for the case of $\alpha_{0}=1e-4,  \beta_{0}=1e-6$ Alpha-MAML converged in less than 15 iterations for a wide range of $\alpha_\text{hyperlr}$ and $\beta_\text{hyperlr}$, whereas MAML takes 165 iterations. Same is the case with $\alpha_{0}=1e-5,  \beta_{0}=1e-7$, for most of the choices of $\alpha_\text{hyperlr}$ and $\beta_\text{hyperlr}$, Alpha-MAML converged in lesser number of iterations than MAML.}
	\label{fig:grid_search2}
	%\vspace{-0.2in}
\end{figure*}
%%%%%%%%%%%%%%%%%%%%%%%%%%%%%%%%%%%%%

\paragraph{Implementation details}
\label{appendix_implement}
The network has four modules with 3$\times$3 convolutions and 64 filters. This is followed by batch normalization, a ReLU activation, and strided convolutions. The final output is fed into a softmax layer. The images are downsampled to 28$\times$28, and the dimensionality of the last hidden layer is 64.

An augmentation scheme similar to original MAML implementation is applied, where images are augmented with 90 degrees rotated images.

%% file: main.bbl
\begin{thebibliography}{18}
\providecommand{\natexlab}[1]{#1}
\providecommand{\url}[1]{\texttt{#1}}
\expandafter\ifx\csname urlstyle\endcsname\relax
  \providecommand{\doi}[1]{doi: #1}\else
  \providecommand{\doi}{doi: \begingroup \urlstyle{rm}\Url}\fi

\bibitem[Antoniou et~al.(2019)Antoniou, Edwards, and Storkey]{antoniou2018how}
Antreas Antoniou, Harrison Edwards, and Amos Storkey.
\newblock How to train your {MAML}.
\newblock In \emph{International Conference on Learning Representations}, 2019.

\bibitem[Baydin et~al.(2018)Baydin, Cornish, Rubio, Schmidt, and
  Wood]{baydin2018hypergradient}
Atılım~Güneş Baydin, Robert Cornish, David~Martínez Rubio, Mark Schmidt,
  and Frank Wood.
\newblock Online learning rate adaptation with hypergradient descent.
\newblock In \emph{Sixth International Conference on Learning Representations
  (ICLR), Vancouver, Canada, April 30 -- May 3, 2018}, 2018.

\bibitem[Behl et~al.(2018)Behl, Najafi, and
  Torr]{DBLP:journals/corr/abs-1812-01397}
Harkirat~Singh Behl, Mohammad Najafi, and Philip H.~S. Torr.
\newblock Meta learning deep visual words for fast video object segmentation.
\newblock \emph{CoRR}, 2018.

\bibitem[Bengio(2000)]{bengio2000gradient}
Y.~Bengio.
\newblock Gradient-based optimization of hyperparameters.
\newblock \emph{Neural Computation}, 12\penalty0 (8):\penalty0 1889--1900,
  2000.
\newblock \doi{10.1162/089976600300015187}.

\bibitem[Bergstra et~al.(2013)Bergstra, Yamins, and Cox]{bergstra2013making}
J.~Bergstra, D.~Yamins, and D.~D. Cox.
\newblock Making a science of model search: Hyperparameter optimization in
  hundreds of dimensions for vision architectures.
\newblock In \emph{International Conference on Machine Learning}, 2013.

\bibitem[Finn et~al.(2017)Finn, Abbeel, and Levine]{finn2017model}
Chelsea Finn, Pieter Abbeel, and Sergey Levine.
\newblock Model-agnostic meta-learning for fast adaptation of deep networks.
\newblock In \emph{Proceedings of the 34th International Conference on Machine
  Learning}, 2017.

\bibitem[Hochreiter et~al.(2001)Hochreiter, Younger, and
  Conwell]{hochreiter2001learning}
Sepp Hochreiter, A~Steven Younger, and Peter~R Conwell.
\newblock Learning to learn using gradient descent.
\newblock In \emph{International Conference on Artificial Neural Networks},
  pages 87--94. Springer, 2001.

\bibitem[Hutter et~al.(2013)Hutter, Hoos, and
  Leyton-Brown]{hutter2013evaluation}
F.~Hutter, H.~Hoos, and K.~Leyton-Brown.
\newblock An evaluation of sequential model-based optimization for expensive
  blackbox functions.
\newblock In \emph{Proceedings of the 15th Annual Conference Companion on
  Genetic and Evolutionary Computation}, pages 1209--1216. ACM, 2013.

\bibitem[Lake et~al.(2011)Lake, Salakhutdinov, Gross, and
  Tenenbaum]{lake2011one}
Brenden Lake, Ruslan Salakhutdinov, Jason Gross, and Joshua Tenenbaum.
\newblock One shot learning of simple visual concepts.
\newblock In \emph{Proceedings of the Annual Meeting of the Cognitive Science
  Society}, volume~33, 2011.

\bibitem[Lake et~al.(2015)Lake, Salakhutdinov, and Tenenbaum]{lake2015human}
Brenden~M Lake, Ruslan Salakhutdinov, and Joshua~B Tenenbaum.
\newblock Human-level concept learning through probabilistic program induction.
\newblock \emph{Science}, 350\penalty0 (6266):\penalty0 1332--1338, 2015.

\bibitem[Li et~al.(2017)Li, Zhou, Chen, and Li]{meta_sgd}
Zhenguo Li, Fengwei Zhou, Fei Chen, and Hang Li.
\newblock Meta-sgd: Learning to learn quickly for few shot learning.
\newblock \emph{CoRR}, 2017.

\bibitem[Maclaurin et~al.(2015)Maclaurin, Duvenaud, and
  Adams]{maclaurin2015gradient}
D.~Maclaurin, D.~K. Duvenaud, and R.~P. Adams.
\newblock Gradient-based hyperparameter optimization through reversible
  learning.
\newblock In \emph{Proceedings of the 32nd International Conference on Machine
  Learning}, pages 2113--2122, 2015.

\bibitem[Ravi and Larochelle(2017)]{ravi2016optimization}
Sachin Ravi and Hugo Larochelle.
\newblock Optimization as a model for few-shot learning.
\newblock In \emph{Fifth International Conference on Learning Representations
  (ICLR)}, 2017.

\bibitem[Schmidhuber(1987)]{schmidhuber1987evolutionary}
J{\"u}rgen Schmidhuber.
\newblock \emph{Evolutionary principles in self-referential learning, or on
  learning how to learn: the meta-meta-... hook}.
\newblock PhD thesis, Technische Universit{\"a}t M{\"u}nchen, 1987.

\bibitem[Snell et~al.(2017)Snell, Swersky, and Zemel]{NIPS2017_6996}
Jake Snell, Kevin Swersky, and Richard Zemel.
\newblock Prototypical networks for few-shot learning.
\newblock In \emph{Advances in Neural Information Processing Systems}. 2017.

\bibitem[Snoek et~al.(2012)Snoek, Larochelle, and Adams]{snoek2012practical}
J.~Snoek, H.~Larochelle, and R.~P. Adams.
\newblock Practical {Bayesian} optimization of machine learning algorithms.
\newblock In \emph{Advances in Neural Information Processing Systems}, pages
  2951--2959, 2012.

\bibitem[Thrun and Pratt(2012)]{thrun2012learning}
Sebastian Thrun and Lorien Pratt.
\newblock \emph{Learning to learn}.
\newblock Springer Science \& Business Media, 2012.

\bibitem[Vinyals et~al.(2016)Vinyals, Blundell, Lillicrap, kavukcuoglu, and
  Wierstra]{NIPS2016_6385}
Oriol Vinyals, Charles Blundell, Tim Lillicrap, koray kavukcuoglu, and Daan
  Wierstra.
\newblock Matching networks for one shot learning.
\newblock In \emph{Advances in Neural Information Processing Systems}. 2016.

\end{thebibliography}
